\documentclass[10pt,twocolumn,letterpaper]{article}
\usepackage[pagenumbers]{cvpr}
\usepackage{fix-cm}
\usepackage{times,graphicx,amsmath,amssymb,booktabs,array,xcolor}
\usepackage{caption}
\usepackage{algorithm}
\usepackage[noend]{algpseudocode}
\algrenewcommand\algorithmicrequire{\textbf{Input:}}
\algrenewcommand\algorithmiccomment[1]{\hfill{\color{black!55}$\triangleright$~#1}}
\usepackage{microtype}
\usepackage{colortbl}
\usepackage{placeins}
\usepackage{multirow,tabularx}
\newcolumntype{Y}{>{\raggedleft\arraybackslash}X}
\definecolor{bpOurs}{RGB}{0,114,178}
\definecolor{rankBest}{HTML}{FF9999}
\definecolor{rankSecond}{HTML}{FFCC99}
\definecolor{rankThird}{HTML}{FFF6AD}
\AtBeginEnvironment{table}{\color{black}\hypersetup{linkcolor=black,citecolor=black,urlcolor=black}}
\AtBeginEnvironment{table*}{\color{black}\hypersetup{linkcolor=black,citecolor=black,urlcolor=black}}
\newcommand{\gain}[1]{{\scriptsize\textcolor{bpOurs}{#1}}}
\newcommand{\gainlabel}{{\scriptsize\textcolor{bpOurs}{\textit{vs.\ best baseline}}}}

\definecolor{bpGray}{RGB}{110,110,110}

\definecolor{cvprblue}{rgb}{0.21,0.49,0.74}
\usepackage[pagebackref,breaklinks,colorlinks,allcolors=cvprblue]{hyperref}
\usepackage[capitalize]{cleveref}
\hypersetup{pdftitle={Go Far, Look Less: Active Multi-Room Reconstruction with Persistent Goals and Selective Views},
  pdfauthor={Jiawei Jiang, Haonan Zhou, Wendi Hu, Jiabei Hu, Yina Xiang, Kaiyue Zhou, Gaoang Wang}}
\title{Go Far, Look Less: Active Multi-Room Reconstruction\\with Persistent Goals and Selective Views}
\author{Jiawei~Jiang\textsuperscript{1}, Haonan~Zhou\textsuperscript{1}, Wendi~Hu\textsuperscript{1}, Jiabei~Hu\textsuperscript{1}, Yina~Xiang\textsuperscript{1}, Kaiyue~Zhou\textsuperscript{2,*}, Gaoang~Wang\textsuperscript{1,*}\\[3pt]
\textsuperscript{1}Zhejiang University\qquad\textsuperscript{2}Chengdu Minto Tech}
\begin{document}
\twocolumn[{
\renewcommand\twocolumn[1][]{#1}
\maketitle
\begin{minipage}{\textwidth}
\centering
\captionsetup{type=figure,belowskip=10pt}
\includegraphics[width=0.97\linewidth]{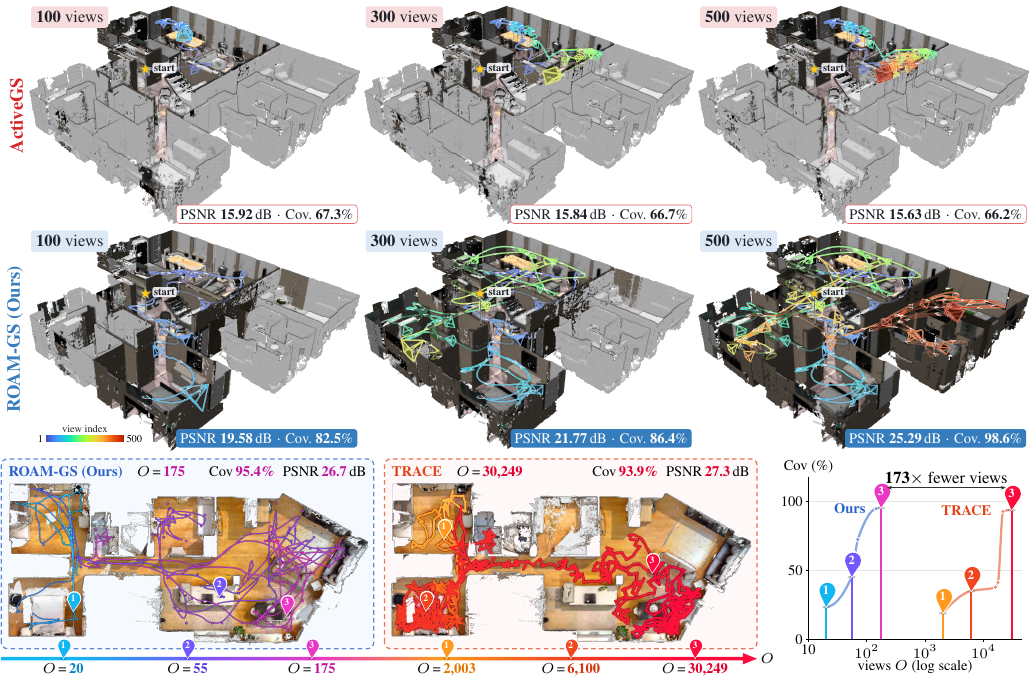}
\caption{\textbf{Go Far, Look Less.} Sparse planners polish nearby rooms, whereas dense mappers reach distant ones only by streaming the trajectory. \textbf{Top:} At equal budgets on HSSD, persistent goals carry ROAM-GS past the local trap of ActiveGS~\cite{activegs} to a complete multi-room reconstruction. \textbf{Bottom:} On HM3D, endpoint-only capture matches TRACE~\cite{trace} mapping with \textbf{173$\times$ fewer views}.}
\label{fig:teaser}
\end{minipage}

}]
\raggedbottom
\begingroup
\renewcommand{\thefootnote}{\fnsymbol{footnote}}
\footnotetext[1]{Corresponding authors.}
\endgroup
\begin{abstract}
Active 3D reconstruction aims to allocate a sparse view budget to recover complete geometry and high-fidelity renderings. However, existing methods degrade sharply in multi-room homes: myopic next-best-view scoring gives transit no credit for distant rooms and over-polishes irreducible nearby residuals, whereas dense mappers reach far rooms only by streaming thousands of frames. We propose \textbf{ROAM-GS}, a training-free active multi-room reconstruction framework that couples persistent goals with selective views: each decision selects a goal and a variable-horizon prefix toward it, capturing only at its endpoint. A Surface Reconstruction Memory scores poses by projecting post-update defects and viewing directions. Persistent goals credit transit with their remaining value and are released once arrivals stop measurably improving them. Without per-scene tuning, ROAM-GS achieves state-of-the-art multi-room reconstruction while preserving single-room Replica fidelity, raising mean coverage from 74\% to \textbf{93\%} on HSSD and from 47\% to \textbf{96\%} (\textbf{+11.7~dB} PSNR) on HM3D, and matching 90-minute dense mapping using \textbf{173$\times$ fewer views}.
\par
\end{abstract}
\noindent Project page: \url{https://roamgs.github.io}.\par

\section{Introduction}
\label{sec:intro}

Active 3D reconstruction lets an embodied agent build geometric and visual models of an unfamiliar scene by choosing where to observe~\cite{feng2024naruto,isler2016information,activegs,activesplat,xu2026area3d}. In multi-room homes connected by doorways and corridors, the agent must allocate a sparse view budget online across the entire house.

Existing methods select views by coverage, uncertainty, or information gain~\cite{activegamer,isler2016information,fisherrf,activegs,li2026activepolicy,popgs,gaussmi,xu2026area3d,zhang2026peering}, learn policies from prior scenes~\cite{chaplot2020ans,chen2024gennbv,guedon2023macarons,li2025nextbestpath}, plan hierarchically~\cite{activesplat,hgsplanner}, or optimize continuous trajectories~\cite{trace}. In multi-room homes, they face a dilemma between view sparsity and house-scale coverage (\cref{fig:teaser}). Sparse view planners are efficient in single rooms but stall across houses, leaving distant rooms unreconstructed. Dense mappers reach those rooms only by streaming frames along the entire trajectory, at the cost of tens of thousands of images.

Sparse planners misallocate views in multi-room homes for two reasons. \textbf{First}, myopic step-wise scoring gives corridor transit no credit for the value of its distant destination and replans at every step, so a robot that has rebuilt its living room keeps polishing the sofa and leaves the next room unvisited (\cref{fig:teaser}, top). \textbf{Second}, residual magnitude overstates what new views can fix, as glossy or finely detailed surfaces keep high rendering errors long after optimization stops improving them. As a result, distant rooms are abandoned too early and nearby defects are pursued too long. How can an agent reconstruct all rooms of a home under a sparse view budget?

\begin{figure*}[t]
  \centering
  \includegraphics[width=\linewidth]{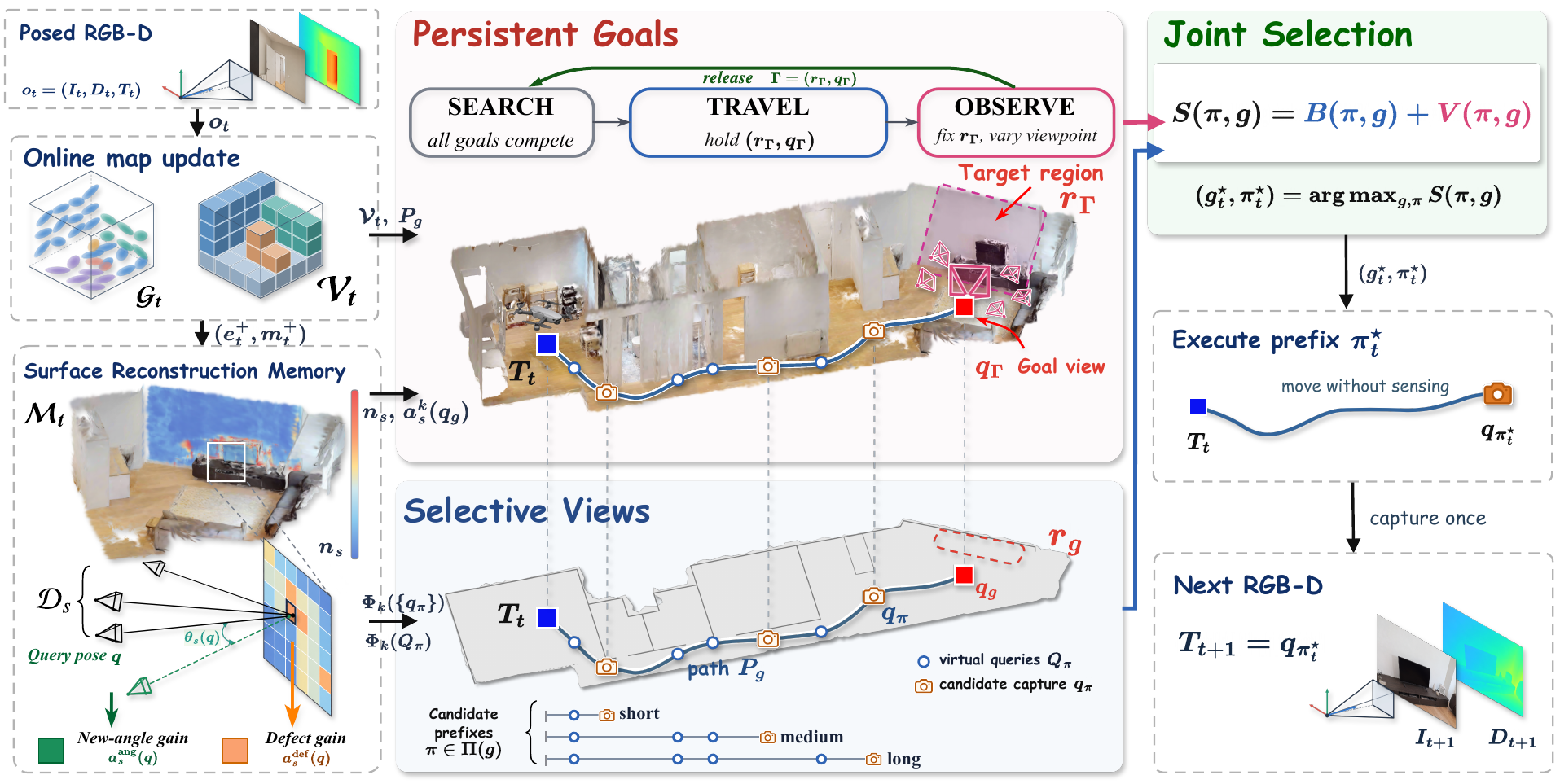}
  \caption{\textbf{Overview of ROAM-GS.} Each posed RGB-D capture updates the Gaussian map $\mathcal{G}_t$, the occupancy grid $\mathcal{V}_t$, and the Surface Reconstruction Memory $\mathcal{M}_t$, whose cells store observed directions $\mathcal{D}_s$ and score query views by new-angle and defect gains without rendering. \emph{Persistent goals} (top) keep a target region $r_g$ through search, travel, and observe until an arrival capture measurably improves it. \emph{Selective views} (bottom) truncate the path to the goal view at variable horizons: only the prefix endpoint (orange) is captured, while query poses (hollow) are credited from memory. The joint score $S=B+V$ selects the goal and prefix, and the agent captures once at $q_\pi$.}
  \label{fig:overview}
\end{figure*}

We propose \textbf{ROAM-GS}, an active reconstruction framework that decouples where the agent heads from where it captures (\cref{fig:overview}). Each decision selects a \emph{goal}, \ie, a target region with a viewpoint that observes it, and a prefix of the path toward it; the agent traverses the prefix without sensing and captures only at its endpoint. A 3D \emph{Surface Reconstruction Memory} records post-update defects and viewing directions on surface cells, so that endpoints, routes, and goals are scored by geometric projection without rendering. \emph{Persistent goals} resolve both causes of misallocation. The selected goal stays active across decisions and its transit prefixes inherit the value it still offers beyond the route, so distant rooms are reached; the goal is released only when an arrival measurably improves its target region, so irreducible residuals stop attracting views. \emph{Selective views} avoid dense streaming: long prefixes cross corridors for a single capture, short prefixes inspect locally, and surfaces passed along the route are credited without being captured.

Our contributions are threefold:
\begin{itemize}
    \item \textbf{Persistent goals with selective views.} We propose a training-free multi-room reconstruction framework that jointly selects a goal and a variable-horizon path prefix under one score, holds the goal until measured improvement releases it, and captures only at prefix endpoints.
    \item \textbf{Surface Reconstruction Memory.} We introduce a 3D memory of post-update defects and viewing directions that scores endpoints, virtual routes, and goals by geometric projection, without Gaussian rendering.
    \item \textbf{State-of-the-art multi-room reconstruction.} Without per-scene tuning, ROAM-GS raises mean coverage from $74\%$ to \textbf{93\%} on HSSD~\cite{hssd} and from $47\%$ to \textbf{96\%} (\textbf{+11.7~dB} PSNR) on HM3D~\cite{hm3d}, matches 90-minute dense mapping using \textbf{173$\times$ fewer views}, and preserves single-room Replica fidelity~\cite{replica}.
\end{itemize}

\section{Related Work}
\label{sec:related}

\noindent\textbf{View Utility for Active Reconstruction.}
Differentiable radiance fields~\cite{nerf,instantngp} and Gaussian representations~\cite{kerbl2023gaussian,gaussian2d,sugar,gaussian_surfels} support online dense mapping~\cite{imap,niceslam,coslam,splatam,monogs,gsslam}. Active reconstruction builds on such maps and scores candidate views by geometric coverage~\cite{connolly1985determination,pcnbv}, volumetric information gain~\cite{isler2016information}, uncertainty~\cite{feng2024naruto,neunbv,pan2022activenerf,neurar,active_implicit_nbv,activer_map,active3d,zhang2026peering}, Gaussian confidence or information gain~\cite{fisherrf,activegs,popgs,gaussmi,activegamer,li2026activepolicy}, or learned and semantic priors~\cite{chen2024gennbv,guedon2022scone,guedon2023macarons,stair,xu2026area3d}. Some methods also account for viewing directions~\cite{activegs} or redundancy between views~\cite{li2026activepolicy,trace}. However, these methods value individual candidate views, whereas a sparse multi-room planner must also value the route toward a view and the distant goal it serves. ROAM-GS records post-update defects and viewing directions in 3D cells, scoring endpoints, routes, and goals from one Surface Reconstruction Memory by geometric projection without Gaussian rendering.

\medskip
\noindent\textbf{Long-Horizon Exploration and Planning.}
Frontier exploration~\cite{yamauchi1997frontier}, receding-horizon and sampling-based planners~\cite{bircher2016receding,informed_sampling,schmid_ipp}, and hierarchical frameworks~\cite{tare,fuel} plan beyond individual views. Active neural and Gaussian mapping systems adopt graph-, sampling-, or hierarchy-based exploration~\cite{yan2023active,anm_at_scale,gsplanner,activesplat,hgsplanner}, and TRACE~\cite{trace} optimizes continuous ergodic trajectories. Learned planners predict long-term goals~\cite{chaplot2020ans,gleam}, and NextBestPath~\cite{li2025nextbestpath} predicts the surface coverage accumulated along paths, trained on prior scenes. Under a sparse view budget, a planner must decide where along the route to capture and when to abandon a goal. ROAM-GS holds goals until measured improvement releases them and captures only at variable-horizon prefix endpoints, without offline policy training.

\section{Method}
\label{sec:method}

\subsection{Overview}
\label{sec:overview}

ROAM-GS reconstructs an unknown multi-room environment online under a budget of $N$ posed RGB-D captures (\cref{fig:overview}).
At step $t$, the agent records a capture $o_t=(I_t,D_t,T_t)$, comprising a color image $I_t$, a depth map $D_t$, and the camera pose $T_t$.
A 3D Gaussian map $\mathcal{G}_t$~\cite{activegs,kerbl2023gaussian} represents the reconstruction, and a voxel occupancy grid $\mathcal{V}_t$ labels space as occupied, free, or unknown and supplies frontiers~\cite{yamauchi1997frontier} and collision-free paths.
A Surface Reconstruction Memory $\mathcal{M}_t$ (\cref{sec:memory}) records which surfaces remain unresolved and from which directions they have been observed, so that a pose can be evaluated before the agent visits it.
Starting from empty states $(\mathcal{G}_0,\mathcal{V}_0,\mathcal{M}_0)$, each capture updates all three incrementally:
\begin{equation}
  (\mathcal{G}_{t-1},\mathcal{V}_{t-1},\mathcal{M}_{t-1}) \xrightarrow{o_t} (\mathcal{G}_t,\mathcal{V}_t,\mathcal{M}_t).
  \label{eq:map_update}
\end{equation}
The agent then chooses the next capture pose $T_{t+1}$, and this loop repeats until all $N$ captures are spent.

\paragraph{Decision space.}
Choosing $T_{t+1}$ involves two coupled questions: where the agent heads, and how far it travels before capturing again.
ROAM-GS answers them with two objects that decouple motion from sensing.
A \emph{goal} $g=(r_g,q_g)$ pairs a target region $r_g$, either a frontier or a cluster of under-reconstructed surface cells, with a viewpoint $q_g$ that observes it; $\mathcal{C}_t$ denotes the candidate goals at step $t$.
A \emph{prefix} $\pi\in\Pi(g)$ is an initial segment of the planned path to $q_g$ (\cref{sec:views}); the agent traverses it without sensing and captures only at its endpoint, so $T_{t+1}=q_\pi$.

\paragraph{One score, two decisions.}
Both choices are made under a single goal-conditioned score $S(\pi,g)=B(\pi,g)+V(\pi,g)$.
The view utility $B$ rewards what the prefix itself observes, and the goal value $V$ rewards what remains to be gained at $q_g$ beyond the prefix:
\begin{equation}
\begin{aligned}
  g_t^\star&=\operatorname*{arg\,max}_{g\in\mathcal{C}_t^{\Gamma}}\;\max_{\pi\in\Pi(g)} S(\pi,g),\\
  \pi_t^\star&=\operatorname*{arg\,max}_{\pi\in\Pi(g_t^\star)} S(\pi,g_t^\star),
  \qquad T_{t+1}=q_{\pi_t^\star}.
\end{aligned}
  \label{eq:decision}
\end{equation}
A goal is thus valued by its best prefix, and the executed prefix is then chosen under the selected goal.
The following subsections build $S$ from the bottom up.
\Cref{sec:memory} defines the per-cell gains from which both terms are computed.
\emph{Persistent goals} (\cref{sec:goals}) define $V$ and the candidate set $\mathcal{C}_t^{\Gamma}\subseteq\mathcal{C}_t$ of the outer maximization: the selected goal is held as the \emph{active goal} $\Gamma$ until measured feedback releases it, and all candidates compete, \ie, $\mathcal{C}_t^{\Gamma}=\mathcal{C}_t$, only when no goal is held.
\emph{Selective views} (\cref{sec:views}) construct the prefixes $\Pi(g)$ of the inner maximization and define $B$.

\subsection{Surface Reconstruction Memory}
\label{sec:memory}

Both terms of $S$ must predict which unresolved surfaces an unvisited pose would see and from which direction.
An uncertainty map rendered in a single image plane cannot answer this, since it is tied to the pose it was rendered from and records no viewing directions.
ROAM-GS therefore stores reconstruction quality on the surfaces themselves, in a memory $\mathcal{M}_t$ of 3D surface cells into which the backprojected depth points of each capture are indexed (\cref{fig:overview}).

\paragraph{Cell statistics.}
Each cell $s$ stores the set $\mathcal{D}_s$ of directions from which it has been observed and running averages $(\bar e_s,\bar m_s)$ of the per-pixel color residual and missing-coverage (low-opacity) indicator.
Let $(e_t^-,m_t^-)$ and $(e_t^+,m_t^+)$ denote these per-pixel quantities rendered at $T_t$ from $\mathcal{G}_{t-1}$ and $\mathcal{G}_t$, \ie, before and after integrating $o_t$.
The averages use only the post-update values, because residuals that survive optimization mark defects a capture could not repair, whereas pre-update residuals mostly reflect content not yet fitted; the pre-update values are used instead to measure improvement (\cref{sec:goals}).
From these statistics, a defect score $\chi_s$ measures how poorly a cell is reconstructed, and a need $n_s$ measures whether it is still worth observing, being high for defective cells and also for cells observed from fewer than two directions.
Cells of high need are clustered into the surface target regions of candidate goals.

\paragraph{Rendering-free gains.}
A query pose $q$ is scored by projecting every cell into it, which yields two per-cell gains:
\begin{equation}
\begin{aligned}
  a_s^{\mathrm{def}}(q)&=w_s(q)\,\chi_s,\\
  a_s^{\mathrm{ang}}(q)&=w_s(q)\,\kappa\big(\theta_s(q)\big)\,n_s,
\end{aligned}
  \label{eq:cell_gain}
\end{equation}
where $w_s(q)$ combines line-of-sight visibility, distance, and recency, $\theta_s(q)$ is the smallest angle between the viewing ray $d_s(q)$ and $\mathcal{D}_s$, and $\kappa$ ramps from zero at an angle $\theta_0$ to one.
The defect gain rewards seeing unrepaired surfaces, and the angular gain rewards seeing them from a new direction.
Both gains require only geometric projection, without sensing or Gaussian rendering, so many hypothetical poses can be scored at every decision.
Goal value $V$ (\cref{sec:goals}) and view utility $B$ (\cref{sec:views}) aggregate these cell gains.

\subsection{Persistent Goals}
\label{sec:goals}

\begin{table*}[t]
\centering
\caption{Quantitative comparison on the HSSD and HM3D datasets under a 500-view budget. Homes are grouped by room count into Small, Medium, and Large, which contain 2, 2, and 1 homes on HSSD and 1, 2, and 1 homes on HM3D. Cells report the mean $\pm$ standard deviation over three seeds, averaged over the homes in each group. The last row of each dataset gives the difference from the best baseline. PSNR in dB, Cov10 in \%, and CM in cm.}
\label{tab:main}
{\fontsize{7.8}{9.0}\selectfont\setlength{\tabcolsep}{1.0pt}\renewcommand{\arraystretch}{1.12}
\newcommand{\mainstat}[2]{{\fontsize{8.0}{9.0}\selectfont #1{\fontsize{5.6}{6.4}\selectfont$\pm$\,#2}}}
\begin{tabularx}{\linewidth}{@{}c@{\hspace{1pt}}l*{9}{Y}@{}}
\toprule
& & \multicolumn{3}{c}{Small} & \multicolumn{3}{c}{Medium} & \multicolumn{3}{c}{Large}\\
\cmidrule(lr){3-5}\cmidrule(lr){6-8}\cmidrule(l){9-11}
& Method & PSNR$\uparrow$ & Cov10$\uparrow$ & CM$\downarrow$ & PSNR$\uparrow$ & Cov10$\uparrow$ & CM$\downarrow$ & PSNR$\uparrow$ & Cov10$\uparrow$ & CM$\downarrow$ \\
\midrule
\multirow{7}{*}{\rotatebox[origin=c]{90}{\textbf{HSSD}}} & FisherRF~\cite{fisherrf} & \mainstat{13.01}{0.34} & \mainstat{21.5}{0.1} & \mainstat{98.3}{0.2} & \cellcolor{rankThird}\mainstat{19.30}{1.46} & \mainstat{55.0}{16.5} & \cellcolor{rankThird}\mainstat{39.1}{16.0} & \mainstat{14.49}{0.03} & \mainstat{6.7}{0.1} & \mainstat{407.6}{0.1} \\[1.0pt]
 & POp-GS~\cite{popgs} & \mainstat{13.70}{0.85} & \mainstat{27.3}{6.7} & \mainstat{80.0}{21.5} & \mainstat{18.67}{0.43} & \mainstat{48.1}{6.3} & \mainstat{43.6}{12.2} & \cellcolor{rankThird}\mainstat{14.60}{0.13} & \mainstat{7.2}{1.1} & \mainstat{404.5}{1.2} \\[1.0pt]
 & NARUTO~\cite{feng2024naruto} & \cellcolor{rankThird}\mainstat{15.97}{1.42} & \cellcolor{rankThird}\mainstat{42.6}{9.0} & \cellcolor{rankThird}\mainstat{49.9}{1.4} & \mainstat{15.12}{0.12} & \mainstat{34.5}{1.8} & \mainstat{92.1}{11.0} & \mainstat{13.82}{0.50} & \cellcolor{rankThird}\mainstat{13.5}{2.3} & \cellcolor{rankSecond}\mainstat{117.9}{72.2} \\[1.0pt]
 & ActiveSplat~\cite{activesplat} & \mainstat{9.95}{0.62} & \mainstat{42.4}{3.3} & \mainstat{56.9}{2.5} & \mainstat{10.37}{0.05} & \cellcolor{rankThird}\mainstat{68.9}{1.4} & \mainstat{41.9}{0.5} & \mainstat{3.39}{0.06} & \mainstat{10.0}{0.8} & \mainstat{350.4}{2.7} \\[1.0pt]
 & ActiveGS~\cite{activegs} & \cellcolor{rankSecond}\mainstat{21.10}{1.94} & \cellcolor{rankSecond}\mainstat{78.9}{7.2} & \cellcolor{rankSecond}\mainstat{28.6}{9.1} & \cellcolor{rankSecond}\mainstat{27.18}{0.18} & \cellcolor{rankSecond}\mainstat{98.2}{0.5} & \cellcolor{rankSecond}\mainstat{7.4}{0.1} & \cellcolor{rankSecond}\mainstat{16.64}{0.75} & \cellcolor{rankSecond}\mainstat{16.4}{4.5} & \cellcolor{rankThird}\mainstat{322.6}{38.4} \\[1.0pt]
 & \textbf{ROAM-GS (ours)} & \cellcolor{rankBest}\mainstat{25.31}{0.14} & \cellcolor{rankBest}\mainstat{97.5}{0.1} & \cellcolor{rankBest}\mainstat{3.8}{0.1} & \cellcolor{rankBest}\mainstat{27.49}{0.41} & \cellcolor{rankBest}\mainstat{98.7}{0.2} & \cellcolor{rankBest}\mainstat{7.2}{$<$0.1} & \cellcolor{rankBest}\mainstat{22.97}{0.38} & \cellcolor{rankBest}\mainstat{74.1}{0.1} & \cellcolor{rankBest}\mainstat{11.7}{4.7} \\[1.0pt]
 & \gainlabel & \gain{+4.2} & \gain{+18.6} & \gain{$-$24.8} & \gain{+0.3} & \gain{+0.5} & \gain{$-$0.2} & \gain{+6.3} & \gain{+57.7} & \gain{$-$106.2} \\
\midrule
\multirow{7}{*}{\rotatebox[origin=c]{90}{\textbf{HM3D}}} & FisherRF~\cite{fisherrf} & \mainstat{9.72}{0.39} & \mainstat{29.9}{1.8} & \mainstat{68.6}{8.9} & \mainstat{9.56}{0.99} & \mainstat{15.7}{3.8} & \mainstat{131.6}{11.8} & \mainstat{7.31}{1.07} & \mainstat{7.4}{3.5} & \mainstat{273.3}{73.1} \\[1.0pt]
 & POp-GS~\cite{popgs} & \mainstat{9.73}{0.54} & \cellcolor{rankThird}\mainstat{32.3}{2.9} & \mainstat{65.4}{7.9} & \mainstat{10.18}{0.49} & \mainstat{20.3}{2.8} & \mainstat{124.2}{2.1} & \cellcolor{rankThird}\mainstat{8.37}{0.47} & \cellcolor{rankThird}\mainstat{12.7}{1.4} & \cellcolor{rankThird}\mainstat{224.1}{18.4} \\[1.0pt]
 & NARUTO~\cite{feng2024naruto} & \mainstat{12.15}{0.47} & \mainstat{22.5}{1.7} & \cellcolor{rankThird}\mainstat{50.6}{5.4} & \cellcolor{rankThird}\mainstat{14.48}{0.60} & \mainstat{21.6}{2.8} & \mainstat{101.1}{11.7} & \cellcolor{rankSecond}\mainstat{9.42}{0.51} & \cellcolor{rankSecond}\mainstat{14.22}{3.21} & \cellcolor{rankSecond}\mainstat{199.12}{45.19} \\[1.0pt]
 & ActiveSplat~\cite{activesplat} & \cellcolor{rankThird}\mainstat{12.76}{0.01} & \mainstat{19.5}{0.1} & \mainstat{162.2}{0.3} & \mainstat{11.06}{0.27} & \cellcolor{rankThird}\mainstat{25.6}{0.8} & \cellcolor{rankThird}\mainstat{99.1}{6.4} & \mainstat{6.69}{0.08} & \mainstat{11.6}{0.7} & \mainstat{234.4}{2.0} \\[1.0pt]
 & ActiveGS~\cite{activegs} & \cellcolor{rankSecond}\mainstat{18.59}{5.28} & \cellcolor{rankSecond}\mainstat{81.0}{23.9} & \cellcolor{rankSecond}\mainstat{22.9}{28.0} & \cellcolor{rankSecond}\mainstat{15.84}{0.51} & \cellcolor{rankSecond}\mainstat{50.7}{3.5} & \cellcolor{rankSecond}\mainstat{63.7}{9.5} & \mainstat{6.42}{0.02} & \mainstat{6.5}{0.5} & \mainstat{327.9}{2.3} \\[1.0pt]
 & \textbf{ROAM-GS (ours)} & \cellcolor{rankBest}\mainstat{27.33}{0.40} & \cellcolor{rankBest}\mainstat{97.2}{0.3} & \cellcolor{rankBest}\mainstat{3.1}{$<$0.1} & \cellcolor{rankBest}\mainstat{26.08}{0.34} & \cellcolor{rankBest}\mainstat{97.4}{1.5} & \cellcolor{rankBest}\mainstat{3.3}{0.7} & \cellcolor{rankBest}\mainstat{23.82}{0.29} & \cellcolor{rankBest}\mainstat{90.5}{2.2} & \cellcolor{rankBest}\mainstat{7.1}{0.4} \\[1.0pt]
 & \gainlabel & \gain{+8.7} & \gain{+16.2} & \gain{$-$19.8} & \gain{+10.2} & \gain{+46.7} & \gain{$-$60.4} & \gain{+14.4} & \gain{+76.3} & \gain{$-$192.0} \\
\bottomrule
\end{tabularx}}
\end{table*}

Captures on the way to an unseen room observe little, so a step-wise planner is repeatedly drawn to nearby residual defects and never reaches distant rooms.
A \emph{persistent goal} credits transit toward a distant goal, holds that goal across decisions, and releases it only when measured feedback shows that further captures stop helping.

\paragraph{Transit inherits goal value.}
For a distant goal to win a decision at all, its prefixes must be credited, before they reach $q_g$, with the gain that $q_g$ still offers beyond what the route already credits:
\begin{equation}
\begin{aligned}
  V(\pi,g) &= \frac{\sum_k\omega_k\Psi_k(g\mid\pi)}{1+L_{\mathrm{rem}}(\pi,g)/L_0},\\
  \Psi_k(g\!\mid\!\pi) &= \min\!\Bigl(1,\eta\nu\!\sum\nolimits_s\bigl[a_s^k(q_g)-\tilde a_s^k(\pi,g)\bigr]_+\Bigr),
\end{aligned}
\label{eq:goal_value}
\end{equation}
where $L_{\mathrm{rem}}(\pi,g)$ is the remaining path length from $q_\pi$ to $q_g$ with scale $L_0$, $\omega_k$ are fixed channel weights, $\eta$ is a normalization constant, $\nu\in(0,1]$ is a region trust weight set by feedback (below), and $[x]_+=\max(0,x)$.
The route credit $\tilde a_s^k(\pi,g)$ is the largest gain that any virtual query pose in $Q_\pi$, sampled along $\pi$ before its endpoint (\cref{sec:views}), offers cell $s$ from a direction within $\theta_0$ of $d_s(q_g)$.
Because the route term of $B$ already credits these cells, subtracting $\tilde a$ counts each cell once within $S$; it does not mark them as observed, so surfaces that $q_g$ sees from a novel direction keep their value.
Besides the surface channels $k\in\{\mathrm{def},\mathrm{ang}\}$, an unknown-space channel $k=\mathrm{unk}$ applies the same route subtraction to the unknown voxels of $\mathcal{V}_t$.
As the agent advances, $L_{\mathrm{rem}}$ shrinks and the credit grows; once $q_\pi=q_g$, the endpoint term of $B$ already scores this content, so we set $V(\pi,g)=0$.

\paragraph{Commitment holds the goal.}
The goal value lets a distant goal win one decision, but a free replan at the next step can still trade it for newly exposed local defects.
ROAM-GS therefore locks the selected goal as the \emph{active goal} $\Gamma=(r_\Gamma,q_\Gamma)\leftarrow g_t^\star$ and restricts the candidate set $\mathcal{C}_t^{\Gamma}$ of \cref{eq:decision} in three stages (\cref{fig:overview}).
In \emph{search}, no goal is active and $\mathcal{C}_t^{\Gamma}=\mathcal{C}_t$.
In \emph{travel}, $\mathcal{C}_t^{\Gamma}=\{\Gamma\}$, so $S$ only selects the horizon of the next prefix toward $q_\Gamma$, which is kept while this path remains feasible.
In \emph{observe}, entered once a prefix reaches $q_\Gamma$, $\mathcal{C}_t^{\Gamma}$ pairs the fixed region $r_\Gamma$ with alternative viewpoints, which are reached by short prefixes and compete to replace $q_\Gamma$.
Thus, the target region persists until release, whereas its viewpoint persists only during travel.

\paragraph{Measured improvement releases the goal.}
Residual magnitude alone is an unreliable stopping signal: glossy or finely detailed surfaces keep high rendering errors long after additional views stop improving them.
ROAM-GS instead tests whether an \emph{arrival capture}, \ie, a capture recorded at $q_\Gamma$, actually improves $r_\Gamma$.
Using the pre- and post-update values of \cref{sec:memory}, it averages their reduction over the pixels $\Omega_t$ that backproject onto $r_\Gamma$:
\begin{equation}
  \Delta e_t=\big\langle e_t^- - e_t^+\big\rangle_{\Omega_t},
  \qquad
  \Delta m_t=\big\langle m_t^- - m_t^+\big\rangle_{\Omega_t}.
  \label{eq:gain}
\end{equation}
An arrival is \emph{productive} if $\Delta e_t\ge\delta_e$ or $\Delta m_t\ge\delta_m$; frontier goals use an analogous occupancy test.
A productive arrival releases $\Gamma$, and $r_\Gamma$ competes again under its updated need.
An unproductive arrival keeps $\Gamma$ so that alternative viewpoints can be tried, and decays $\nu$ in \cref{eq:goal_value}; after $N_f$ consecutive unproductive arrivals, $\Gamma$ is released and $r_\Gamma$ is temporarily suspended.
Because only arrival captures are tested, transit captures never penalize a distant goal.

\subsection{Selective Views}
\label{sec:views}

Along the route to a fixed goal, consecutive frames observe heavily overlapping content, much of it already reconstructed, so capturing continuously or after every short motion spends the view budget on redundant frames.
\emph{Selective views} instead choose how far to travel before each capture and credit uncaptured content along the route.

\paragraph{Variable-horizon prefixes.}
Let $P_g$ be the collision-free waypoint path from $T_t$ to $q_g$ planned on $\mathcal{V}_t$.
Truncating it at a set of horizons $\mathcal{H}$ gives the prefixes of \cref{eq:decision}, $\Pi(g)=\{P_g[1{:}\min(h,|P_g|)]\}_{h\in\mathcal{H}}$.
A long prefix crosses an entire corridor or room for the cost of one capture, whereas a short prefix keeps the agent close by for local inspection.
Since the prefixes of every goal in $\mathcal{C}_t^{\Gamma}$ compete under the same score, capture spacing adapts to the scene geometry without a hand-crafted switch between exploration and inspection.

\paragraph{Endpoint and route utility.}
The view utility combines the content captured at the endpoint, the content passed along the route, and a motion term:
\begin{equation}
\begin{aligned}
  B(\pi,g) &= \boldsymbol{\alpha}^{\!\top}\boldsymbol{\Phi}(\{q_\pi\})
  +\boldsymbol{\beta}^{\!\top}\boldsymbol{\Phi}(Q_\pi)-C(\pi,g),\\
  \Phi_k(Q) &= \min\!\Bigl(1,\eta\!\sum\nolimits_s\max_{q\in Q}a_s^k(q)\Bigr).
\end{aligned}
\label{eq:view}
\end{equation}
Here, $\boldsymbol{\Phi}=(\Phi_{\mathrm{fr}},\Phi_{\mathrm{def}},\Phi_{\mathrm{ang}})\in[0,1]^3$ aggregates the gains of \cref{eq:cell_gain} visible from a pose set $Q$, and the frontier channel takes the same union over the frontier voxels of $\mathcal{V}_t$.
The maximum over $q\in Q$ credits each cell at most once, however many poses see it.
The endpoint term scores the only frame that is actually captured, whereas the route term scores the virtual query poses $Q_\pi\subset\pi\setminus\{q_\pi\}$, which are never captured and are evaluated from $\mathcal{M}_t$ alone.
The route term favors prefixes that pass near unresolved surfaces, because the next decision starts from $q_\pi$ and can return to inspect them at close range; separate weights $\boldsymbol{\alpha}$ and $\boldsymbol{\beta}$ value captured and merely passed content differently.
Finally, $C(\pi,g)$ penalizes travel distance, cumulative rotation, and proximity to recent capture poses, and rewards progress toward $q_g$.
With $B$ and $V$ defined, \cref{eq:decision} selects each prefix and its next capture, which updates the maps and memory (\cref{eq:map_update}).

\section{Experiments}
\label{sec:exp}

\begin{figure*}[t]
\centering
\includegraphics[width=\linewidth]{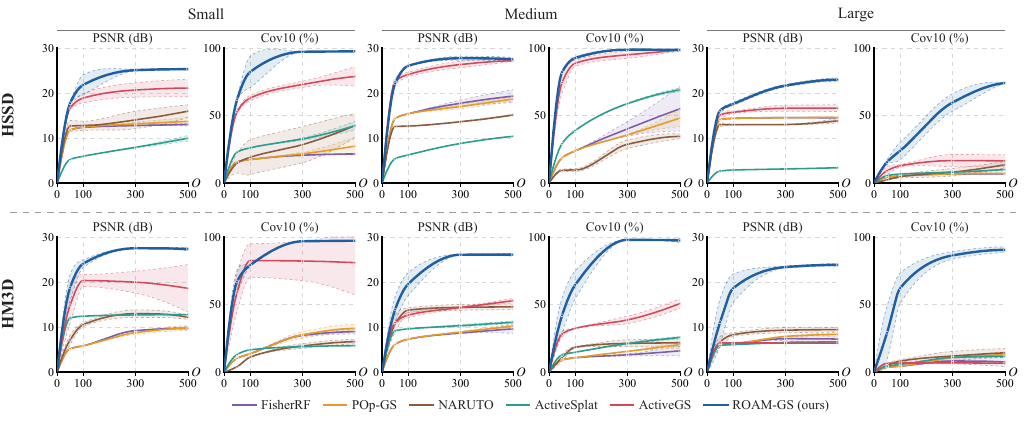}
\caption{PSNR and Cov10 versus captured views on HSSD (top) and HM3D (bottom) for the runs in Table~\ref{tab:main}. Within each Small, Medium, and Large group, PSNR is shown on the left and Cov10 on the right. Curves show means over three seeds, and shaded bands show the standard deviation at 50, 100, 300, and 500 views.}
\label{fig:progression}
\end{figure*}

\begin{figure*}[t]
\centering
\includegraphics[width=\linewidth]{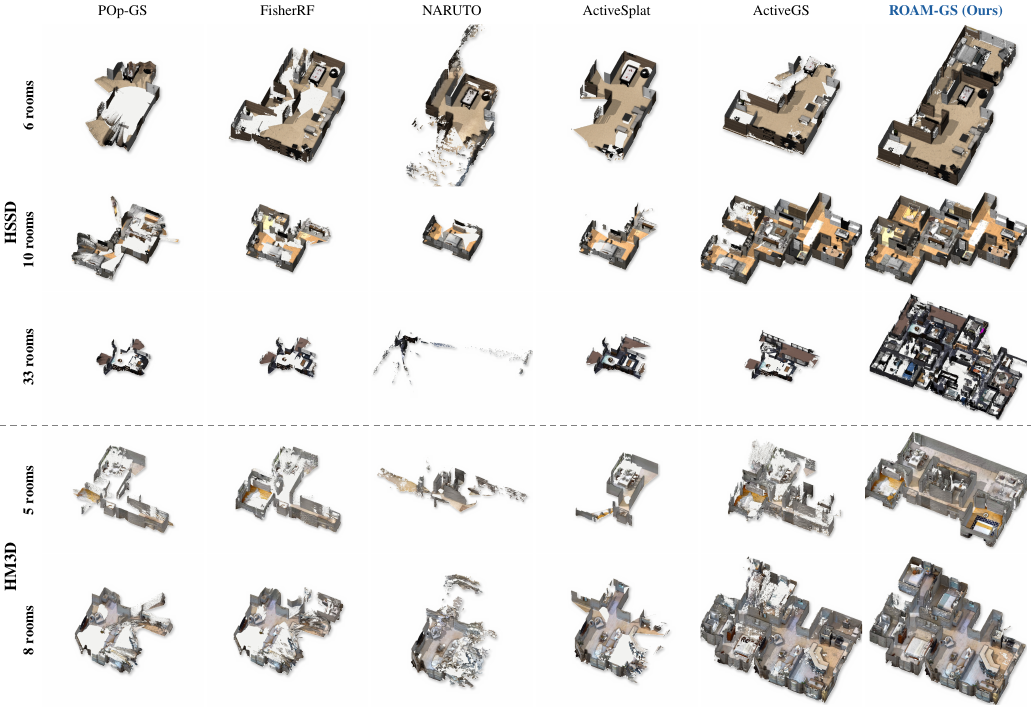}
\caption{Qualitative comparison on HSSD and HM3D after 500 views. ROAM-GS avoids local trapping and achieves substantially higher multi-room coverage.}
\label{fig:qualitative}
\end{figure*}

\begin{figure*}[tp]
\centering
\includegraphics[width=\linewidth]{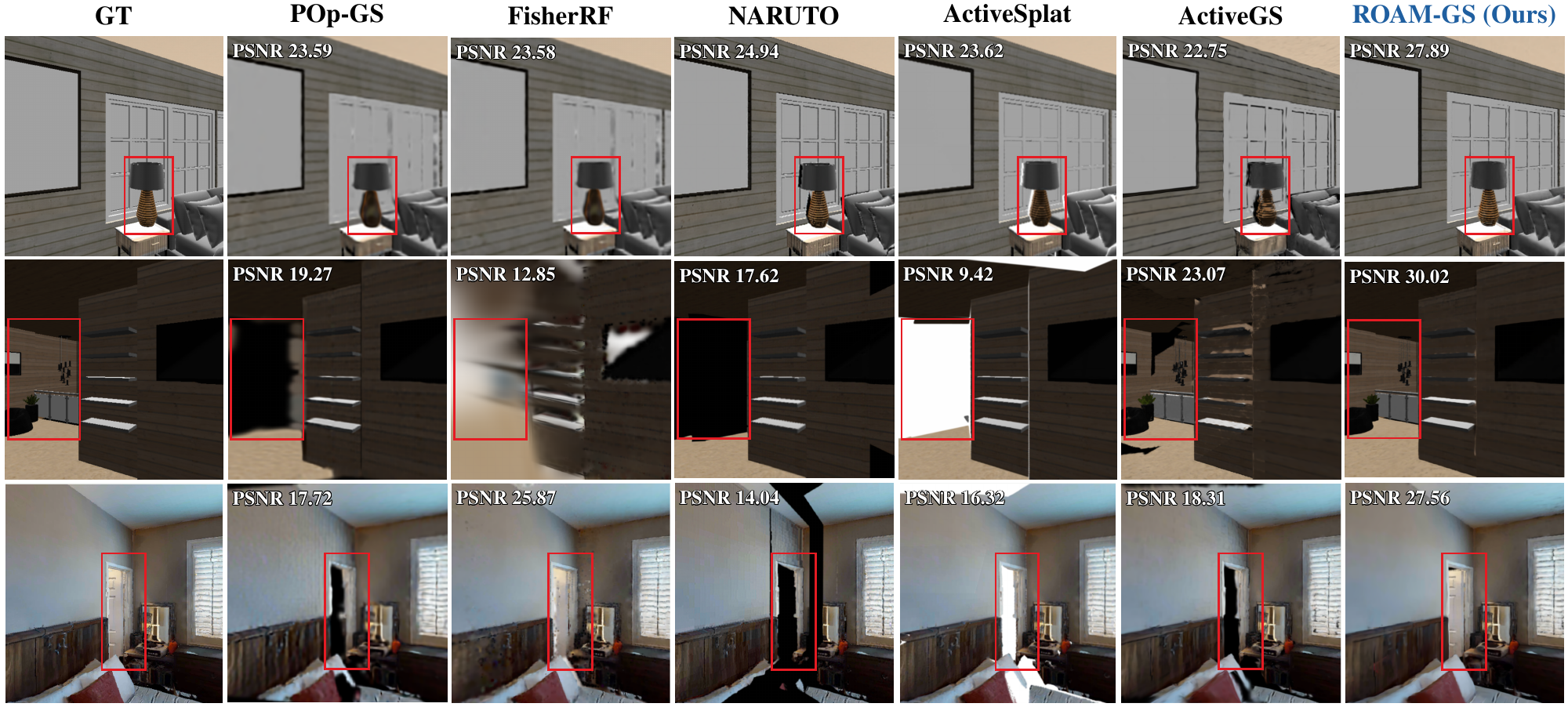}
\caption{Novel view synthesis on HSSD and HM3D after 500 views. Test viewpoints are excluded from the trajectories of all methods. PSNR (dB) is shown in the top-left corner of each view, and red boxes highlight challenging regions.}
\label{fig:rgb_comparison}
\end{figure*}

\subsection{Implementation Details}
\label{sec:exp_setup}

\paragraph{Platform and sensor.}
We implement ROAM-GS in the Habitat simulator~\cite{habitat}. The agent controls its 3D position, yaw, and pitch with fixed roll, and each capture uses one pitch angle selected from $\{-60^\circ,0^\circ,60^\circ\}$. The RGB-D sensor has a $60^\circ \times 60^\circ$ field of view and retains depth measurements in $D\in[0.1,5]\,\mathrm{m}$. Following ActiveGS~\cite{activegs}, the mapper runs 10 optimizer updates per capture, and ROAM-GS integrates every capture into $\mathcal{G}_t$. All experiments run on a single NVIDIA RTX 4090 GPU.

\begin{table}[!b]
\centering
\captionsetup{width=\linewidth,justification=raggedright,singlelinecheck=false}
\begingroup
\centering
\small
\setlength{\tabcolsep}{2.2pt}
\renewcommand{\arraystretch}{1.08}
\caption{Native wall-clock comparison on a 5-room HM3D home. $O$, $B$, and $U$ denote captured views, admitted mapping frames, and optimizer updates. Highlights indicate the top three results across all rows.}
\label{tab:native90}
\resizebox{\linewidth}{!}{\begin{tabular}{@{}lrrrrrrr@{}}
\toprule
\textbf{Method} & \textbf{Min.} & $O \downarrow$ & $B \downarrow$ & $U$ & \textbf{Path (m)} & \textbf{Cov10} $\uparrow$ & \textbf{PSNR} $\uparrow$ \\
\midrule
\multicolumn{8}{@{}l}{\emph{Baselines at 60 min}} \\
ActiveGS~\cite{activegs} & 60.0 & 2,018 & 2,018 & 20,180 & 746.7 & 16.98 & 6.64 \\
ActiveSplat~\cite{activesplat} & 60.0 & 1,703 & 1,703 & 1,203 & \cellcolor{rankBest}29.7 & 77.80 & 19.14 \\
TRACE~\cite{trace} & 60.0 & 17,679 & 4,223 & 5,088 & 230.0 & 41.20 & 11.82 \\
\addlinespace[2pt]
\multicolumn{8}{@{}l}{\emph{Baselines at 90 min}} \\
ActiveGS~\cite{activegs} & 90.0 & 2,652 & 2,652 & 26,520 & 978.8 & 16.91 & 6.59 \\
ActiveSplat~\cite{activesplat} & 90.0 & 2,103 & 2,103 & 1,559 & \cellcolor{rankSecond}31.6 & 82.38 & 20.52 \\
TRACE~\cite{trace} & 90.0 & 30,249 & 8,273 & 9,948 & 464.0 & 93.9 & 27.3 \\
\midrule
\multicolumn{8}{@{}l}{\emph{ROAM-GS (Ours) progression across wall-clock checkpoints}} \\
\textbf{ROAM-GS (10 min)} & 10.1 & 94 & 94 & 940 & \cellcolor{rankThird}174.9 & 83.75 & 24.17 \\
\textbf{ROAM-GS (20 min)} & 20.0 & 183 & 183 & 1,830 & 337.7 & \cellcolor{rankThird}98.25 & \cellcolor{rankThird}28.16 \\
\textbf{ROAM-GS (60 min)} & 60.0 & 569 & 569 & 5,690 & 554.79 & \cellcolor{rankSecond}99.05 & \cellcolor{rankSecond}29.51 \\
\textbf{ROAM-GS (90 min)} & 90.1 & 828 & 828 & 8,280 & 654.8 & \cellcolor{rankBest}99.38 & \cellcolor{rankBest}29.81 \\
\bottomrule
\end{tabular}}
\endgroup

\end{table}

\newcommand{\replicacomparisontable}{
\begin{table}[t]
\centering
\captionsetup{width=\linewidth,justification=raggedright,singlelinecheck=false}
\begingroup
\centering
\caption{Quantitative comparison on Replica Office0, Office1, and Room0 with a matched budget of 200 mapping frames. Values are averaged over the three scenes. PSNR in dB, Cov10 and Cov5 in \%, and CM in cm.}
\label{tab:single}
{\fontsize{7.0}{8.5}\selectfont\setlength{\tabcolsep}{1.8pt}\renewcommand{\arraystretch}{1.05}
\begin{tabularx}{\linewidth}{@{}l*{5}{>{\centering\arraybackslash}X}>{\columncolor{white}[\tabcolsep][0pt]\centering\arraybackslash}X@{}}
\toprule
\textbf{Method} & \textbf{PSNR}$\uparrow$ & \textbf{SSIM}$\uparrow$ & \textbf{LPIPS}$\downarrow$ & \textbf{Cov10}$\uparrow$ & \textbf{Cov5}$\uparrow$ & \textbf{CM}$\downarrow$ \\
\midrule
FisherRF~\cite{fisherrf} & 30.87 & 0.9051 & 0.2293 & 87.01 & 82.27 & 3.771 \\
POp-GS~\cite{popgs} & 27.51 & 0.8375 & 0.2634 & 67.22 & 63.69 & 10.183 \\
NARUTO~\cite{feng2024naruto} & 29.22 & 0.8728 & 0.3126 & \cellcolor{rankThird}99.43 & \cellcolor{rankThird}97.71 & 2.241 \\
ActiveSplat~\cite{activesplat} & 21.89 & 0.8833 & 0.1889 & 84.16 & 80.89 & 5.858 \\
TRACE~\cite{trace} & \cellcolor{rankThird}34.33 & \cellcolor{rankThird}0.9482 & \cellcolor{rankSecond}0.0730 & 98.20 & 96.09 & \cellcolor{rankBest}2.050 \\
ActiveGS~\cite{activegs} & \cellcolor{rankSecond}36.32 & \cellcolor{rankSecond}0.9588 & \cellcolor{rankBest}0.0727 & \cellcolor{rankSecond}99.60 & \cellcolor{rankSecond}98.92 & \cellcolor{rankThird}2.108 \\
\textbf{ROAM-GS} & \cellcolor{rankBest}36.75 & \cellcolor{rankBest}0.9604 & \cellcolor{rankThird}0.0763 & \cellcolor{rankBest}99.94 & \cellcolor{rankBest}99.24 & \cellcolor{rankSecond}2.076 \\
\bottomrule
\end{tabularx}}
\endgroup

\end{table}
}

\paragraph{Datasets.}
We evaluate multi-room reconstruction on HSSD~\cite{hssd} (five synthetic homes with $6$--$33$ rooms) and HM3D~\cite{hm3d} (four real scans with $5$--$15$ rooms). By room count we group homes into Small, Medium, and Large ($1$--$9$, $10$--$20$, $\ge 21$ on HSSD; $\le 5$, $6$--$9$, $\ge 10$ on HM3D). All active planners are given a budget of 500 views on HSSD and HM3D. Single-room fidelity is tested on three Replica scenes~\cite{replica}: Office0, Office1, and Room0.

\paragraph{Metrics.}
We report PSNR on held-out views rendered from 1,000 uniformly distributed test viewpoints on Replica and 3,000 on HSSD and HM3D, completeness as the reference surface covered within $10\text{ cm}$ and $5\text{ cm}$ (Cov10, Cov5), and accuracy as symmetric Chamfer distance (CM, cm). We additionally report SSIM and LPIPS on Replica. Acquisition cost is measured by captured images, mapped frames, optimizer updates, path length, and wall-clock time.

\paragraph{Baselines.}
We compare against five active planners: FisherRF~\cite{fisherrf}, POp-GS~\cite{popgs}, NARUTO~\cite{feng2024naruto}, ActiveSplat~\cite{activesplat}, and ActiveGS~\cite{activegs}, and the dense mapper TRACE~\cite{trace}, which streams frames under its native schedule (\cref{sec:exp_dense}). All methods run with their native policies from the same start poses, using the same simulator and sensor. The top-three results are highlighted ({\setlength{\fboxsep}{1.5pt}\colorbox{rankBest}{1st}, \colorbox{rankSecond}{2nd}, \colorbox{rankThird}{3rd}}).

\subsection{Results on HSSD \& HM3D}
\label{sec:exp_multi}

ROAM-GS achieves the best mean performance in every room group of both datasets (Table~\ref{tab:main}). Averaged over all homes, it raises Cov10 over the strongest baseline, ActiveGS, from 74.1\% to 93.3\% on HSSD and from 47.2\% to 95.6\% on HM3D, while PSNR gains 3.1\,dB and 11.7\,dB and Chamfer distance falls $11.7\times$ and $26.9\times$ against the best baselines. Fig.~\ref{fig:progression} plots PSNR and Cov10 against captured views.

\begin{table}[t]
\centering
\captionsetup{width=\linewidth,justification=raggedright,singlelinecheck=false}
\begingroup
\centering
\caption{Quantitative comparison on Replica Office0, Office1, and Room0 with a matched budget of 200 mapping frames. Values are averaged over the three scenes. PSNR in dB, Cov10 and Cov5 in \%, and CM in cm.}
\label{tab:single}
{\fontsize{7.0}{8.5}\selectfont\setlength{\tabcolsep}{1.8pt}\renewcommand{\arraystretch}{1.05}
\begin{tabularx}{\linewidth}{@{}l*{5}{>{\centering\arraybackslash}X}>{\columncolor{white}[\tabcolsep][0pt]\centering\arraybackslash}X@{}}
\toprule
\textbf{Method} & \textbf{PSNR}$\uparrow$ & \textbf{SSIM}$\uparrow$ & \textbf{LPIPS}$\downarrow$ & \textbf{Cov10}$\uparrow$ & \textbf{Cov5}$\uparrow$ & \textbf{CM}$\downarrow$ \\
\midrule
FisherRF~\cite{fisherrf} & 30.87 & 0.9051 & 0.2293 & 87.01 & 82.27 & 3.771 \\
POp-GS~\cite{popgs} & 27.51 & 0.8375 & 0.2634 & 67.22 & 63.69 & 10.183 \\
NARUTO~\cite{feng2024naruto} & 29.22 & 0.8728 & 0.3126 & \cellcolor{rankThird}99.43 & \cellcolor{rankThird}97.71 & 2.241 \\
ActiveSplat~\cite{activesplat} & 21.89 & 0.8833 & 0.1889 & 84.16 & 80.89 & 5.858 \\
TRACE~\cite{trace} & \cellcolor{rankThird}34.33 & \cellcolor{rankThird}0.9482 & \cellcolor{rankSecond}0.0730 & 98.20 & 96.09 & \cellcolor{rankBest}2.050 \\
ActiveGS~\cite{activegs} & \cellcolor{rankSecond}36.32 & \cellcolor{rankSecond}0.9588 & \cellcolor{rankBest}0.0727 & \cellcolor{rankSecond}99.60 & \cellcolor{rankSecond}98.92 & \cellcolor{rankThird}2.108 \\
\textbf{ROAM-GS} & \cellcolor{rankBest}36.75 & \cellcolor{rankBest}0.9604 & \cellcolor{rankThird}0.0763 & \cellcolor{rankBest}99.94 & \cellcolor{rankBest}99.24 & \cellcolor{rankSecond}2.076 \\
\bottomrule
\end{tabularx}}
\endgroup

\end{table}

\paragraph{The largest gains are observed in the Large groups.}
In the Large group every baseline covers under 20\% of the reference surface, while ROAM-GS covers 74.1\% of the 33-room HSSD house and 90.5\% of the 15-room HM3D home. As seen in Fig.~\ref{fig:qualitative}, next-best-view planners finish the rooms near the start and leave the rest empty, whereas persistent goals carry the agent to distant rooms. Fig.~\ref{fig:rgb_comparison} compares rendered RGB views and local reconstruction details.

\paragraph{Comparison with dense mapping.}
\label{sec:dense_time}
\label{sec:exp_dense}
Under native 90-minute schedules on a 5-room HM3D home (Table~\ref{tab:native90}), ROAM-GS reaches 83.75\% coverage in 10 minutes and surpasses 90-minute TRACE within 20 (98.25\% vs.\ 93.9\% Cov10 and 28.16 vs.\ 27.3\,dB PSNR) using $165\times$ fewer images and $45\times$ fewer mapped frames; after 90 minutes it further refines to 99.38\% coverage and 29.81\,dB PSNR.

\subsection{Results on Replica}
\label{sec:exp_single}

On Replica with a matched budget of 200 mapping frames, ROAM-GS reaches 36.75\,dB PSNR, 99.94\% Cov10, and 2.076\,cm CM (Table~\ref{tab:single}). It leads PSNR, SSIM, Cov10, and Cov5, exceeding ActiveGS by 0.43\,dB PSNR and 0.34 Cov10 points, while ActiveGS achieves the lowest LPIPS and TRACE the lowest CM. These results show that multi-room gains coexist with high single-room fidelity.

\subsection{Ablation Study}
\label{sec:exp_ablation}

We ablate each component on the HSSD and HM3D datasets under a 500-view budget from a common initial pose, keeping the rest of the pipeline unchanged (Table~\ref{tab:ablation}).

\begin{table}[htbp]
\centering
\small
\setlength{\tabcolsep}{1.8pt}
\renewcommand{\arraystretch}{1.00}
\setlength{\aboverulesep}{0.35ex}
\setlength{\belowrulesep}{0.50ex}
\caption{Ablation on HSSD and HM3D under a 500-view budget. Best is bold. PSNR in dB, Cov10 in \%, and CM in cm.}
\label{tab:ablation}
\label{tab:ablation_main}
\resizebox{\linewidth}{!}{\begin{tabular}{@{}lcccccc@{}}
\toprule
 & \multicolumn{3}{c}{\textbf{HSSD}} & \multicolumn{3}{c}{\textbf{HM3D}} \\
\cmidrule(lr){2-4} \cmidrule(lr){5-7}
\textbf{Variant} & \textbf{PSNR}$\uparrow$ & $\textbf{Cov10}\uparrow$ & \textbf{CM}$\downarrow$ & \textbf{PSNR}$\uparrow$ & $\textbf{Cov10}\uparrow$ & \textbf{CM}$\downarrow$ \\
\midrule
\textbf{Full ROAM-GS} & \textbf{24.91} & \textbf{96.8} & \textbf{3.04} & \textbf{27.08} & \textbf{97.5} & 2.51 \\
\midrule
\multicolumn{7}{@{}l}{\emph{Surface Reconstruction Memory}} \\
w/o memory $\mathcal{M}_t$ & 17.72 & 71.4 & 26.32 & 21.45 & 78.6 & 18.40 \\
w/o route credit $\tilde a_s^k$ & 22.57 & 90.7 & 4.71 & 25.14 & 91.9 & 4.48 \\
\midrule
\multicolumn{7}{@{}l}{\emph{Persistent goals}} \\
w/o goal value $V(\pi,g)$ & 23.54 & 94.7 & 4.20 & 25.46 & 95.2 & \textbf{2.40} \\
w/o task commitment $\Gamma$ & 20.66 & 79.0 & 22.90 & 22.84 & 83.2 & 14.75 \\
w/o measured release & 23.18 & 91.2 & 5.45 & 25.30 & 92.8 & 4.18 \\
\midrule
\multicolumn{7}{@{}l}{\emph{Selective views}} \\
Stepwise capture ($h = 1$) & 18.38 & 54.0 & 40.36 & 19.90 & 88.6 & 20.87 \\
Goal-only capture ($h = |P_g|$) & 22.70 & 93.1 & 4.05 & 24.85 & 92.4 & 4.62 \\
\bottomrule
\end{tabular}}
\end{table}

\paragraph{Surface Reconstruction Memory.}
The memory supplies the per-cell gains from which both the goal value and the view utility are computed (\cref{sec:memory}). Removing $\mathcal{M}_t$ lowers Cov10 by 25.4 and 18.9 points and PSNR by 7.19 and 5.63\,dB on HSSD and HM3D, and raises Chamfer distance by $8.7\times$ and $7.3\times$. Removing the route credit $\tilde a_s^k$ lowers Cov10 by 6.1 and 5.6 points, more than removing the goal value $V(\pi,g)$ entirely (2.1 and 2.3 points), so double-counted transit credit hurts more than no transit credit. Each surface must be credited once across the route and goal.

\paragraph{Persistent goals.}
Task commitment is the most critical goal component. Removing $\Gamma$ lowers Cov10 by 17.8 and 14.3 points and PSNR by over 4.2\,dB, and raises Chamfer distance by $7.5\times$ and $5.9\times$, as step-wise replanning abandons distant goals mid-transit. Disabling measured release lowers Cov10 by 5.6 and 4.7 points and PSNR by over 1.7\,dB, showing that before-and-after gains are needed to decide when to leave a region. Removing goal value $V(\pi,g)$ costs 1.37 and 1.62\,dB PSNR and 2.1 and 2.3 Cov10 points, while HM3D Chamfer distance stays similar (2.40 vs.\ 2.51), so lookahead credit mainly improves completeness and appearance.

\paragraph{Selective views.}
Both extremes of the prefix horizon $h$ (\cref{sec:views}) underperform the adaptive set $\mathcal{H}$. Stepwise capture ($h=1$) loses 6.53 and 7.18\,dB PSNR, and lowers Cov10 by 42.8 and 8.9 points, as short strides exhaust the budget before reaching distant rooms. Goal-only capture ($h=|P_g|$) skips intermediate observations, losing over 2.2\,dB PSNR and 3.7--5.1 Cov10 points. Adaptive prefixes thus balance transit stride with local inspection.

\section{Conclusion}
\label{sec:conclusion}

We presented ROAM-GS, a training-free active reconstruction framework for multi-room homes that decouples where the agent heads from where it captures. A Surface Reconstruction Memory records post-update defects and viewing directions on 3D surface cells and scores endpoints, routes, and goals by geometric projection without rendering. Persistent goals credit transit with the value of distant destinations and are released only when arrivals measurably improve their target regions, while selective views capture only at the endpoints of variable-horizon path prefixes. Without per-scene tuning, ROAM-GS achieves state-of-the-art multi-room completeness and rendering quality on HSSD and HM3D, preserves single-room Replica fidelity, and surpasses 90-minute dense mapping within 20 minutes using $165\times$ fewer images.

\FloatBarrier
\section{Limitations}
\label{sec:limitations}
All evaluated scenes lie on a single floor. ROAM-GS plans paths in three dimensions, but our experiments leave open how well it can maintain a goal on another floor while navigating stairs.

Our experiments also assume a static layout. If a door closes after the agent passes through it, the occupancy grid may still mark the doorway as traversable. A moved object can likewise leave outdated geometry and viewing history in the map. Handling these changes would require detecting the affected regions and updating the Gaussian map, the occupancy grid, and the Surface Reconstruction Memory together.
\section{LLM Usage}
We used AI assistants in manuscript preparation and experimentation. Claude primarily assisted with drafting and refining the manuscript, while AI coding assistants supported code development and experimental workflows. The research methodology and overall framework were conceived and designed by the authors, who take responsibility for the scientific claims, experimental results, and final manuscript.

{\small
\bibliographystyle{ieeenat_fullname}
\bibliography{main}}

\end{document}